\documentclass[]{spie}  

\usepackage{amsmath,amsfonts,amssymb}
\usepackage{graphicx}
\usepackage{subfigure}
\usepackage{booktabs}

\title{Semantic Segmentation of Panoramic Images Using a Synthetic Dataset}
\author{Yuanyou Xu, Kaiwei Wang, Kailun Yang, Dongming Sun and Jia Fu}
\affil{College of Optical Science and Engineering, Zhejiang University, 310027 Hangzhou, China}

\authorinfo{Correspondence: wangkaiwei@zju.edu.cn}

\begin{document}
\maketitle

\begin{abstract}
Panoramic images have advantages in information capacity and scene stability due to their large field of view (FoV). In this paper, we propose a method to synthesize a new dataset of panoramic image. We managed to stitch the images taken from different directions into panoramic images, together with their labeled images, to yield the panoramic semantic segmentation dataset denominated as SYNTHIA-PANO. For the purpose of finding out the effect of using panoramic images as training dataset, we designed and performed a comprehensive set of experiments. Experimental results show that using panoramic images as training data is beneficial to the segmentation result. In addition, it has been shown that by using panoramic images with a 180 degree FoV as training data the model has better performance. Furthermore, the model trained with panoramic images also has a better capacity to resist the image distortion. Our codes and SYNTHIA-PANO dataset are available: \emph{https://github.com/Francis515/SYNTHIA-PANO}
\end{abstract}


\section{INTRODUCTION}
\label{sec:intro}  

With the development of deep learning, the research on image analysis methods has been boosted. Semantic segmentation, different from the target detection technology, can extract information in the image at the pixel level~\cite{yang2018unifying}\cite{yang2018unifyingterrainawareness}. Currently, most semantic segmentation researches are based on images with a conventional field of view (FoV). The images with a conventional FoV that is relatively narrow can only cover information in a certain direction, and their content will change with respect to the viewpoint, so they have shortcomings in the aspect of information capacity and stability. In contrast, 360-degree panoramic images can compensate for these shortcomings. However, due to their large FoV and distortion, the general semantic segmentation method and training data are not ideal for the segmentation of panoramic images~\cite{yang2019can}. The accuracy will decrease and the general input size of a model requires cutting the panoramic image into several segments~\cite{yang2019can}, which will lead to discontinuity in the segmentation map. In order to realize the better segmentation of panoramic images, we created a dataset of panoramic images and apply it for the training of convolutional neural network to yield a panoramic semantic segmentation model.

The typical method to deal with the distortion in panoramic images is to use data augmentation~\cite{yang2019can}\cite{yang2019robustifying}. The data augmentation can simulate the distortion in panoramic images. When trained with such augmented data, the model can adapt well to panoramic images with distortion. However, the general public datasets for segmentation only contain images of forward-direction view, which can't simulate the large FoV of panoramic images. Our method is to synthesize a new dataset of panoramic images. Our panoramic dataset is built from the virtual image dataset SYNTHIA~\cite{Ros_2016_CVPR} due to the lack of real-world panoramic image dataset. SYHTHIA dateset contains finely labeled images with a conventional FoV. We managed to stitch the images taken from different directions into panoramic images, together with their labeled images, to yield the panoramic semantic segmentation dataset dubbed SYNTHIA-PANO. As panoramic images are generally large in size, we take ICNet~\cite{zhao2018icnet} as the basic model in our work. It is designed for real-time semantic segmentation of images with a large size, which can achieve a good balance between segmentation accuracy and efficiency.

For the purpose of finding out the effect of using panoramic images as training dataset, we designed and performed a comprehensive set of experiments. 
First, we compared different views for training to figure out how this factor affects the performance of the model. Second, we tried images with different FoVs for training and found out the best FoV which can achieve the highest accuracy. Third, images with different degrees of distortion are used to evaluate the anti-distortion ability of the model trained with panoramic images with respect to original images. Experimental results show that 
the view of training data is critical for the segmentation of panoramic images and using panoramic images with a 180-degree FoV for training will obtain the model with better performance. In addition, the model trained with panoramic images has a better capacity to resist the image distortion.

\section{RELATED WORK}

As for the environment perception techniques, the robustness and comprehensiveness of surrounding sensing has become a new focus of attention. Omnidirectional vision sensors can indeed obtain a wider FoV than traditional pinhole cameras. However, recent works are mainly paying more attention to visual localization/odometry~\cite{8461042}\cite{cheng2019panoramic}\cite{chen2019palvo} and monocular depth estimation~\cite{song2018im2pano3d}\cite{Garanderie_2018_ECCV}\cite{tateno2018distortion}\cite{sharma2019unsupervised} when using omnidirectional cameras. Instead, panoramic semantic segmentation, which is capable of compensating for partial shortcomings, has not been clearly studied. L. Deng et al.\cite{7995725} put forward a CNN-based semantic segmentation solution whose to achieve a larger FoV by introducing the fisheye camera. Synchronously, Overlapping Pyramid Pooling (OPP)\cite{Zhao_2017_CVPR} module is applied to effectively deal with the distortion of fisheye images. On this basis, they built a panoramic view system by utilizing four fisheye cameras and proposed Restricted Deformable Convolution (RDC) for semantic segmentation which shows the decent effectiveness in handling images with large distortions~\cite{deng2018restricted}.

W. Zhou et al.\cite{8603770} stitched semantic images via a lens array which contains three 100-degree FoV lens with varying orientations to attain semantic understanding of the wider FoV, even so they still only achieve the 180-degree semantic perception of the forward-view environment. R.~Varga et al.\cite{8317846} used four fisheye cameras, four 360-degree LIDARs and a GPS/IMU sensor to set a super-sensor which is able to complete 360-degree surroundings understanding when having this kind of super-sensor fitted to an automatic vehicle. However, they sacrificed partial vertical FoV for the purpose of preserving straight lines when segmenting unwarpped images on cylindrical projection surfaces. Analogously, K. Narioka et al.\cite{8500397} equipped a vehicle with five cameras equiangularly for a wider FoV to perceive surroundings. They designed a set of efficient Deep Neural Networks (DNNs) and trained only with front facing camera images. As a consequence, they obtained the comparatively accurate results of semantic segmentation and depth estimation.

As is known to all, visual data is essential for CNN-based semantic segmentation as semantic features aredirectly learned via the training procedure. As a result, various datasets have been created for many different purposes. For semantic road scene parsing, Cityscapes\cite{Cordts_2016_CVPR} and Mapillary Vistas\cite{Neuhold_2017_ICCV} are backup options. Cityscapes is composed of 5000 images with superb pixel-level annotations and 20 000 additional images with rough annotations which are captured from 50 different European cities using forward-looking traditional cameras. Vistas is also a large-scale dataset and it has $5\times$ more than the total amount of satisfactory annotated images for Cityscapes. Different from Cityscapes, Vistas contains the images taken from various viewpoints and at diverse conditions regarding weather, season and daytime. However, it's worth noting that data annotation and data collection are time-consuming and costly. Thus, another increasingly popular way to overcome the lack of large-scale datasets is explored by using synthetic data such as SYNTHIA\cite{Ros_2016_CVPR}. The technique of creating SYNTHIA is using the Unity development platform to generate and render virtual cities and automatically acquiring realistic synthetic images with pixel-level annotations. SYNTHIA-Seqs, a complementary set of images, are taken from four 100-degree FoV cameras with certain overlapping which can be utilized to generate panoramic images. Very recently, C. Zhang et al.~\cite{zhang2019orientation} created an Omni-SYNTHIA Version of SYNTHIA to validate their method to perform orientation-aware CNN operations on spherical images by using an icosahedron mesh. Comparatively, in this paper, we offered a PANO-SYNTHIA version of SYNTHIA to study 360-degree semantic segmentation.

Taking the limited amount of data into account, data augmentation is currently used to increase the diversity of training samples, improve the model robustness and avoid overfitting. E. Romera et al.~\cite{8500561} proposed that the extant domain gap can be dramatically decayed by applying appropriate data augmentation methods regarding geometry and texture, which also resulted in better calibrated models. Furthermore, they bridged the gap between daytime and nighttime imagery domain to achieve robust semantic segmentation at night~\cite{romera2019bridging}. R.Varga et al.\cite{8317846} suggested that the distortions of images are unavoidable when using fisheye camaras with wide FoV. To address this trouble, W. Zhou et al.\cite{8603770} explored and evaluated the use of skew and gamma correction in data augmentation techniques and L. Deng et al.\cite{7995725} proposed a zooming policy applied for the images captured by fisheye cameras.

In previous works, we have studied real-time semantic segmentation~\cite{yang2018unifying}\cite{yang2019can}, RGB-D semantic segmentation~\cite{yang2019robustifying}~\cite{hu2019acnet} and importance-aware semantic segmentation~\cite{xiang2019importance}\cite{xiang2019comparative}. By contrast, in this paper we focus on panoramic image semantic segmentation by using a synthetic dataset. Our comprehensive variety of experiments, offers the hints that accurate semantic segmentation for 360-degree panoramic images is achievable to support surrounding sensing.

\section{METHOD}

In this section, the methods we use to do segmentation on panoramic images are stated. Our work is based on the CNN model called ICNet~\cite{zhao2018icnet} and a dataset called SYNTHIA~\cite{Ros_2016_CVPR}. We introduce them first. Then our key work is to make a new dataset consisting of panoramic images, which is also discussed in this section.

\subsection{ICNet and SYNTHIA Dataset}
\label{sec:icnet}
Using convolutional neural network (CNN) is an effective method to do semantic segmentation.  Although a deep neural network can achieve good accuracy, sometimes it suffers from the complexity. Panoramic images are often large in size due to their large FoV, which may cause a high computational consumption if a deep neural network is used. So if a model needs to do fast segmentation on panoramic images, it shouldn't be so heavy or the inference will be too slow. Time efficiency of the neural network model is a key criterion when it comes to panoramic images.

However, for a model there is a tradeoff between the complexity and the performance. A complex model tends to have good performance but high consumption. The Image Cascade Network (ICNet)~\cite{zhao2018icnet} is designed for real-time semantic segmentation of high-resolution images and it balances the accuracy and time consumption well. So it's suitable to do segmentation with ICNet on panoramic images. The network can extract features from different resolutions of inputs and the features are combined by the Cascade Feature Fusion (CFF)~\cite{zhao2018icnet} method.

\begin{figure} [ht]
   \begin{center}
   \begin{tabular}{c} 
   \includegraphics[width=0.75\textwidth]{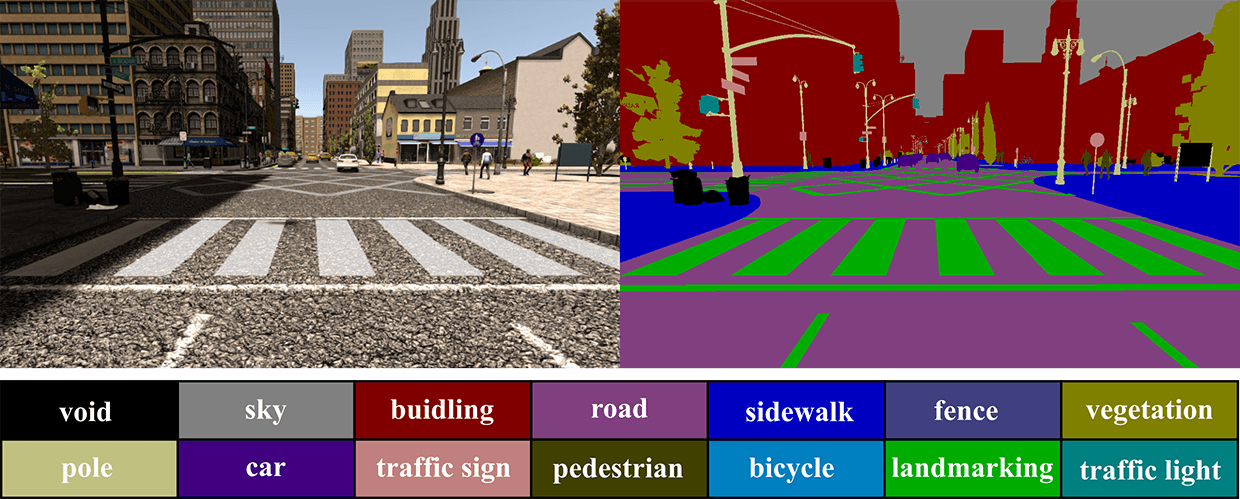}
   \end{tabular}
   \end{center}
   \caption[example]
   { \label{fig:label}
The figure is a demonstration of the the images and labels in SYNTHIA dataset. The color of each class is below.}
   \end{figure}

We choose to use ICNet as the basic model directly and focus our work on the training data. For data-driven CNN models, the training data can have an impact influence on the result. Instead of making structural refinement on the CNN model, we made a new dataset which consists of panoramic images. The new dataset we made is based on another dataset called SYNTHIA. SYNTHIA dataset is created from computer 3D city traffic scene models and all of the images in it are virtual images. It contains some subsets called SYNTHIA-Seqs in which the images are taken by four cameras in leftward, forward, rightward and backward directions on a moving car in the virtual cities. In addition, there are images of different city scenes, seasons, weather conditions and so on. The labels contains 16 classes (2 of the classes are reserved) as it shown in Fig.~\ref{fig:label}. What we did is to stitch these four-direction images into panoramic images. The following sections are about the stitching process.

\subsection{Cylindrical Projection}

One way to get panoramic images is to take images from different directions around a circle and then stitch them together. When the camera rotates, the geometrical relations between the objects in the images also changes. To unify the geometrical relations of the whole scene, cylindrical projection\cite{szeliski1997creating} is an important step before stitching a panoramic image. If the scene is projected on a cylindrical surface, one object in the images from different view directions can be quite the same. In this sense, when stitching the images, the overlapping parts can coincide with each other perfectly.

\begin{figure} [ht]
   \begin{center}
   \begin{tabular}{c} 
   \includegraphics[width=0.4\textwidth]{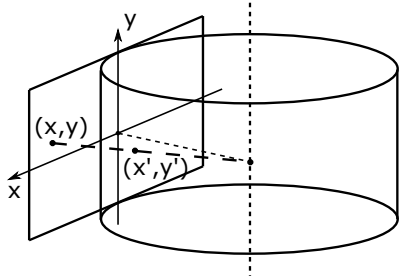}
   \end{tabular}
   \end{center}
   \caption[example]
   { \label{fig:cylinder_prj}
The figure is a demonstration of the cylindrical projection. Every point $(x,y)$ on the image have a corresponding point $(x',y')$ on the surface of the cylinder. The corresponding point is on the line which passes the center point of the cylinder and the original point.}
   \end{figure}

Cylindrical projection is to project an image from a plane to a cylindrical surface, as it shown in Fig.~\ref{fig:cylinder_prj}. Assume that a rectangular image is on a vertical plane $I$, and next to it there is a cylindrical surface $C$ with the radius $r$. The focal length $f$ is the distance between the center of the image and the center of the cylinder. It's common to set $r$ to be equal to the focal length $f$ of the camera. If so, the cylindrical surface $C$ will be tangent to the plane $I$. We establish a Cartesian coordinate system taking the center of the image as the origin. After that, we can build a mapping $F{\cdot}$ from the image plane $I$ to the cylinder surface $C$ . Given a point $p: (x,y)\in I$, the corresponding point $p': (x',y')=F\{(x,y)\}\in C$ can be calculated by the forward mapping equations in Eq.~ \ref{eq:forward_prj}. In the practical implementation, the backward mapping in Eq.~\ref{eq:backward_prj} is more common to use because it can avoid blank pixel appearing in the result.

\begin{equation}
\label{eq:forward_prj}
F\{(x,y);r,f\}=\left\{
\begin{aligned}
x' & = & r\tan^{-1}(\frac{x}{f}) \\
y' & = & \frac{ry}{\sqrt{x^2+f^2}} \\
\end{aligned}
\right.
\end{equation}

\begin{equation}
\label{eq:backward_prj}
F^{-1}\{(x',y');r,f\}=\left\{
\begin{aligned}
x & = & f\tan^{-1}(\frac{x'}{r}) \\
y & = & \frac{y'}{r}\sqrt{x^2+f^2} \\
\end{aligned}
\right.
\end{equation}

The mapping built above is just cylindrical projection and the result of it is an image on cylindrical surface. The most important parameter when doing it is the radius of the cylindrical surface $r$ which is often set the same as the focal length $f$. Generally, a small focal length $f$ leads to severe distortion and vice versa. In our later experiments, we explore different options of focal lengths, and study the influence on the final panoramic segmentation performance.

\subsection{Stitching Method for Images in SYNTHIA Dataset}

\begin{figure} [ht]
   \begin{center}
   \begin{tabular}{c} 
   \includegraphics[width=0.95\textwidth]{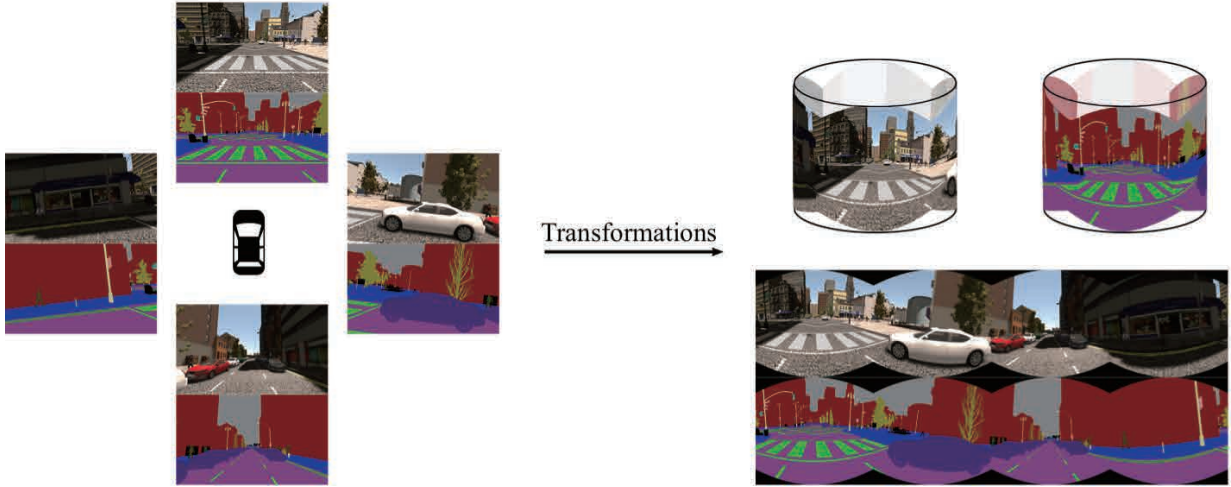}
   \end{tabular}
   \end{center}
   \caption[example]
   { \label{fig:circle}
The figure is a demonstration of the transformations in Eq.~\ref{eq:stitch}. The four images and their labels in leftward, forward, rightward and backward directions are shown on the left of the figure. After the transforms, they are projected on a cylindrical surface and stitched together.}
   \end{figure}

For four images in leftward, forward, rightward and backward directions denoted as $X_f, X_l, X_b, X_r$, we can project the normal images into a cylindrical surface as it is stated in last subsection, and the next step is to stitch them together. Considering that the images are taken by a set of fixed cameras, a panoramic image can be synthesized by a set of transformations denoted as $T\{\cdot\}$. Because once the cameras are fixed the images they take have certain spatial relations and will not change with time or the direction of view. To obtain the panoramic image $X_p$, the stitching method can be described by Eq.~\ref{eq:stitch}.

\begin{equation}
\label{eq:stitch}
X_p=T\{X_l,X_f,X_r,X_b\}=A\cdot F\{X_l\}\cup B\cdot F\{X_f\}\cup C\cdot F\{X_r\}\cup D\cdot F\{X_b\}
\end{equation}
where $A,B,C,D$ are translation transformation matrices and $F\{\cdot\}$ is cylindrical projection. Note that the focal length has already been provide by SYNTHIA Dataset, which is set to be 532.740352. In this regard, if we set $r=f$, the cylindrical projection $F\{\cdot\}$  is then totally determined. The transformations are illustrated in Fig.\ref{fig:circle}.

\begin{figure} [ht]
   \begin{center}
   \begin{tabular}{c} 
   \includegraphics[width=0.7\textwidth]{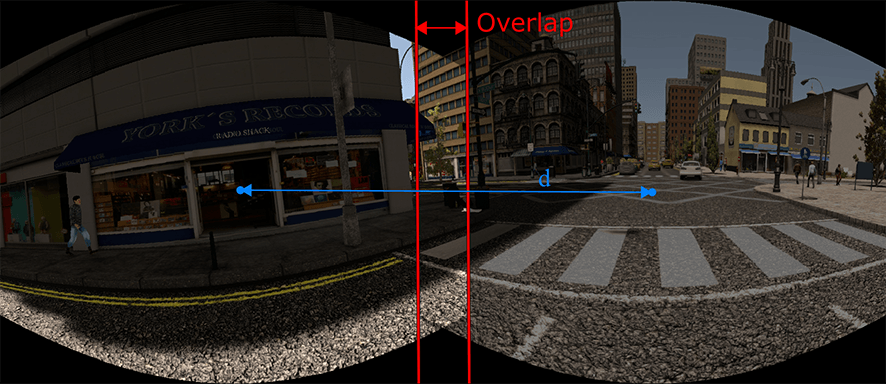}
   \end{tabular}
   \end{center}
   \caption[example]
   { \label{fig:overlap}
The figure is a demonstration of the distance parameter $d$.}
   \end{figure}

The translation transformation\cite{szeliski2007image} from point $(x,y)$ to point $(x',y')$ is defined as Eq.~\ref{eq:translation}.

\begin{equation}
\label{eq:translation}
 \left[
 \begin{matrix}
   x' \\
   y' \\
   1
  \end{matrix}
  \right]
=
 \left[
 \begin{matrix}
   1 & 0 & t_x \\
   0 & 1 & t_y \\
   0 &0 & 1
  \end{matrix}
  \right]
  \cdot
   \left[
 \begin{matrix}
   x \\
   y \\
   1
  \end{matrix}
  \right]
\end{equation}
where the homogeneous coordinates are used. Since the four images are taken at the same height, the shift transformation only contains horizontal component, that is, the items $t_y$ in the shift transformations $A,B,C,D$ which are denoted as $t_{yA},t_{yB},t_{yC},t_{yD}$ are all equal to 0 here. What's more, we can also obtain that

\begin{equation}
\label{eq:adjacent}
t_{xB}-t_{xA}=t_{xC}-t_{xB}=t_{xD}-t_{xC}
\end{equation}
This is because the angle between any two adjacent cameras is the same, i.e., 90 degree. If we set $t_{xA}=0$, then the shift terms will be

\begin{equation}
\label{eq:forward_prj}
\left\{
\begin{aligned}
t_{xA}& = &0 \\
t_{xB}& = &d \\
t_{xC}& = &2d \\
t_{xD}& = &3d \\
\end{aligned}
\right.
\end{equation}
where $d$ is a distance parameter, which means the distance between two adjacent images. This is shown in Fig.~\ref{fig:overlap}

\begin{figure} [ht]
\centering
\subfigure[Result of ORB feature detection and matching] { \label{fig:orb}
\includegraphics[width=0.75\textwidth]{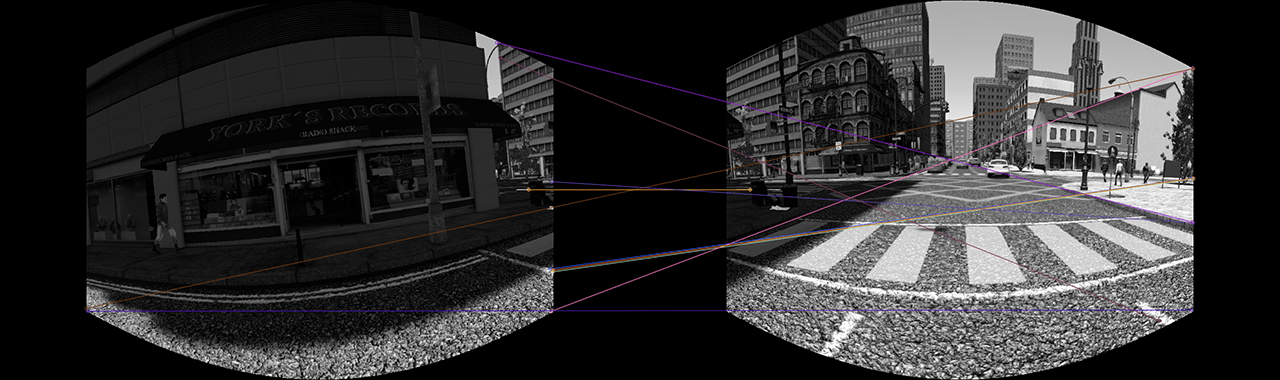}
}

\subfigure[Demonstration of proposed region matching method] { \label{fig:match}
\includegraphics[width=0.75\textwidth]{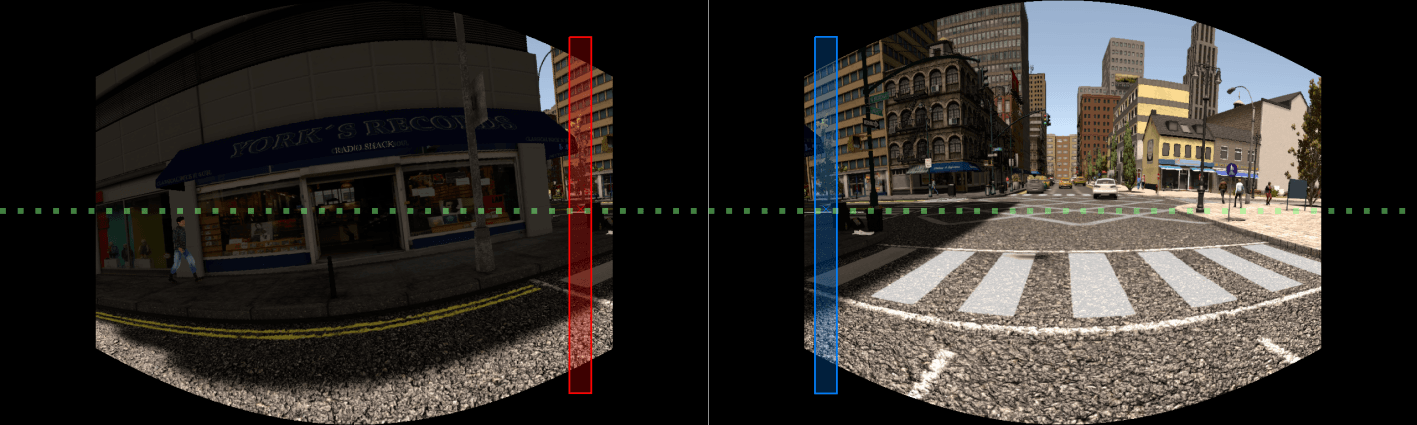}
}
\caption{These two figures show the way to determine the distance parameter $d$: (a) shows the result of ORB feature detection and matching. It can be observed that few features are matched correctly because the overlapping area is thin. (b) shows the result of region matching to find two corresponding regions.}
\label{fig:view}
\end{figure}

In order to stitch four images together, it's necessary to determine the parameter $d$. The traditional method is based on feature detection and matching, by finding two matched feature points and then calculating the parameter $d$ from the coordinates from them. It may not be proper to use the traditional feature detection and matching method to stitching the images in SYNTHIA dataset. A key prerequisite of the traditional method is that the coincident area of two adjacent images for stitching should be large enough. Every image in SYNTHIA Dataset has a horizontal FoV of 100 degree, which means that each of the four direction images around a circle has just approximately 10 degree coincident area. It's not enough for detecting and matching the features. If many of the features are not matched correctly, selecting features that matched correctly will be difficult, as it is illustrated in Fig.~\ref{fig:orb}. When using the ORB~\cite{rublee2011orb} descriptor to detect features and then matching them, the result is not ideal. Only few feature points can be matched correctly. It's not certain that how many feature points can be matched and which pairs are desired.

In view of the above reasons, a specialized method denominated as region matching is proposed in this paper. The method is also like feature-based, but it uses the pixels in the images directly to do matching. It is illustrated in Fig.~\ref{fig:match}. We choose the leftward image $I_1$ and the forward image $I_2$ for example. First, pick out one region in the leftward image, as the red region $R_1\in I_1$ in Fig.~\ref{fig:match}. Then we use the region to scan along the horizontal axis (the green dotted line in Fig.~\ref{fig:match}) in the middle of the forward image from left to right. Each time during scanning there is a corresponding region in the forward image denoted as $R_2\in I_2$, as the blue region in Fig.~\ref{fig:match}. A discrepancy value $Dv$ between the two regions $R_1$ and $R_2$ is defined in Eq.~\ref{eq:dist}.

\begin{equation}
\label{eq:dist}
Dv(R_1,R_2)=\sum_{p\in R_1,q\in R_2}{|p-q|}
\end{equation}
where $p,q$ are pixels in region $R_1,R_2$, respectively. It does not matter whether $R_1$ and $R_2$ is grey or RGB as long as they have the same depth and size. The discrepancy value is negatively related to the similarity of the two regions. A low discrepancy value indicates that the two regions are similar to each other. So as long as we find such two corresponding regions with minimal discrepancy value, we will be able to calculate the distance parameter $d$ from the coordinates of the centers of the two regions. That is similar to what is done in the feature matching.

Let the column coordinates of the centers of $R_1,R_2$ be $x_{c1},x_{c2}$. The $x_{c1}$ is set to be a constant ($x_{c1}=1075$ in the experiment) and $x_{c2}$ is the variable. The region $R_1$ is chosen randomly but it should be in the overlapping region of the two images. So it is proper to pick out a few columns, for example, 9 columns around the edge of the image. Thereby, the aim is to find a $x_{c2}$ which minimizes Eq.~\ref{eq:dist} as Eq.~\ref{eq:min} shows.

\begin{equation}
\label{eq:min}
\underset{x_{c2}}{\arg\min} \: Dv(R_1, R_2)
\end{equation}

         \begin{figure} [ht]
   \begin{center}
   \begin{tabular}{c} 
   \includegraphics[width=0.6\textwidth]{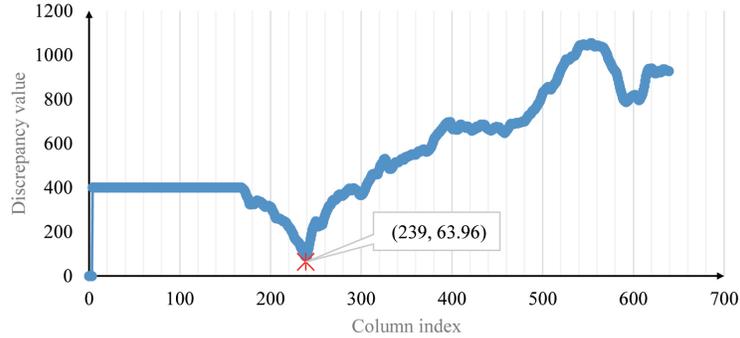}
   \end{tabular}
   \end{center}
   \caption[example]
   { \label{fig:dist_res}
The figure shows the discrepancy value of each column index. The column index of the minimal value can be used to calculate the distance parameter $d$.}
   \end{figure}

 The discrepancy values varies as the $x_{c2}$ increases and the relation is shown in Fig.~\ref{fig:dist_res}. There is an obvious minimal point where $x_{c2}=239$. Then the distance parameter $d$ can be calculated by Eq.~\ref{eq:d}

\begin{equation}
\label{eq:d}
d=x_{c2}-x_{c1}
\end{equation}
When performing the stitching and image cropping, numerical approximation of the cylindrical projection and index counting may introduce 1 or 2 pixel deviation of $d$. The result can be $d\thickapprox835$. To the general case, if the cameras are fixed, the method described above can be applied. Since the traditional feature-based method has relative high complexity, it requires a higher computational capacity. However, the method described above is simple enough and the consumption of time is mainly on the cylindrical projection but not the stitching.

Another issue of panoramic stitching is the brightness inconsistency problem. If the images are taken from different angles, the brightness might be different for the same region in the images. This is not a critical problem in SYNTHIA dataset so we directly perform the stitching without using any other operation. In practical and real conditions, some methods may need to be used to address the brightness inconsistency issue.

\subsection{SYNTHIA-PANO Panoramic Image Dataset }

\begin{figure} [ht]
   \begin{center}
   \begin{tabular}{c} 
   \includegraphics[width=0.95\textwidth]{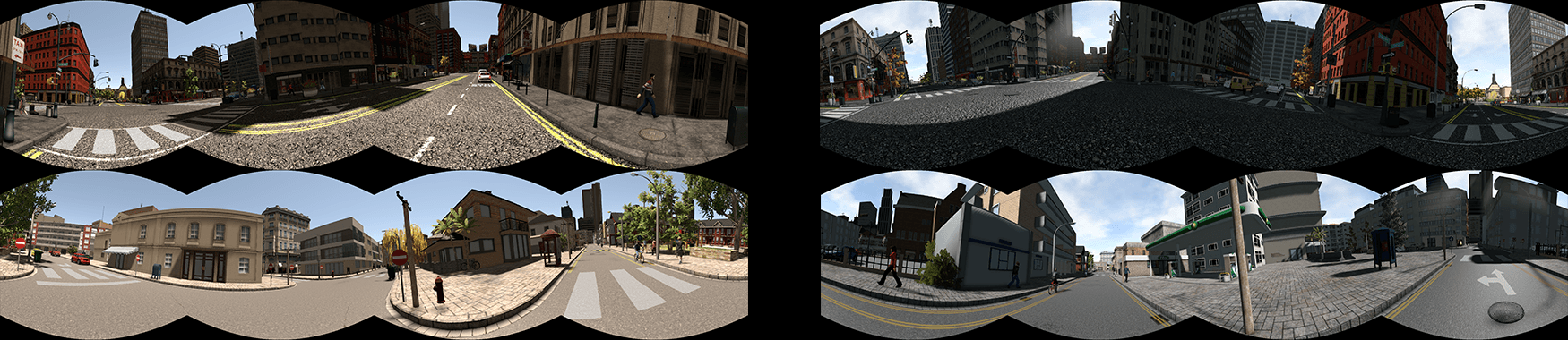}
   \end{tabular}
   \end{center}
   \caption[example]
   { \label{fig:seqs}
The figure shows the panoramic images included in the sequences of SYNTHIA-PANO dataset. The two figures in the first row are Seqs02-summer and Seqs02-fall. The two figures in the second row are Seqs04-summer and Seqs04-fall. Seqs05-summer is similar to Seqs02-summer.}
   \end{figure}

By means of the method above, a panoramic image dataset can be obtained from the original SYNTHIA dataset. Our panoramic image dataset includes five sequences of images: Seqs02-summer, Seqs02-fall, Seqs04-summer, Seqs04-fall, Seqs05-summer. Seqs02 series and Seqs05 series are taken in a New York like city and Seqs04 series are taken in a European town like city. Some examples are shown in Fig.~\ref{fig:seqs}. The original images are ordered video sequences and we truncated the repeated images. The numbers of images in these sequences are displayed in Table.~\ref{tab:synthia_nums}.

When stitching the original images, the order of the directions is randomly chosen so that for any part of a panoramic image it can belong to any direction. Each image in our dataset has the size $(3340, 760)$ and the FoV of 360 degree.

\begin{table}[ht]
\centering
\caption{Information of the sequences in SYNTHIA-PANO dataset}
\label{tab:synthia_nums}
\begin{tabular}{@{}llcc@{}}
\toprule
Sequence name & Scene content      & \multicolumn{1}{l}{Full number of images} & \multicolumn{1}{l}{Number after truncated} \\ \midrule
Seqs02-fall   & New-York like city & 742                                       & 461                                        \\
Seqs02-summer & New-York like city & 888                                       & 556                                        \\
Seqs04-fall   & European town      & 911                                       & 738                                        \\
Seqs04-summer & European town      & 901                                       & 694                                        \\
Seqs05-summer & New-York like city & 787                                       & 787                                        \\ \bottomrule
\end{tabular}
\end{table}

\section{EXPERIMENTS}

In the previous section, a new panoramic image dataset SYNTHIA-PANO is introduced. Based on SYNTHIA-PANO and SYNTHIA dataset, some experiments are designed and performed. First we find out the different views which influence the performance of the model on panoramic images. Then we aim to find out the best FoV to train a CNN model and infer on the panoramic dataset. At last the result of training is evaluated from the aspect of the anti-distortion capacity.

\subsection{Setup for Training}

As for training, a NVIDIA GTX1080Ti GPU is used in following experiments. The CNN model  we use is ICNet which has been introduced in Sec.~\ref{sec:icnet}. The optimizer we use is Adam~\cite{kingma2014adam}, with parameter $\epsilon=1e-7, \beta_1=9e-1, \beta_2=9.99e-1$. The initial learning rate is set to $1e-3$ and learning rate decay policy is used. The whole number of training iterations is set to be a constant while the batch size depends on the specific number of images in the training data. The specific settings of training will be shown in following subsections. The evaluation is based on mainly two indices, the pixel accuracy and the mean intersection over union (mIoU).

For the panoramic images in SYNTHIA-PANO, there are big differences among the numbers of pixels for different classes. For instance, the sky, building and road take a large percent of the whole area in an image and the classes such as traffic sign and traffic light only take a small percent of area. Due to the unbalanced numbers of pixels, a seemingly small loss can't indicate a good result of the true segmentation accuracy. In addition, it will also affect the process of optimization. So the class balancing method is used during training. We counted the pixel number of each class and divided each by the median as the class weight. A class with a small number of pixels will have a larger class weight and vice versa.

\subsection{Different Views for Training}
\label{sec:view}

For inferring on panoramic images, there are mainly two possible reasons that may get the performance of the model worse. The first one is the view direction of the training data, and another one is the distortion of the panoramic images. This subsection is mainly about the view direction factor and the distortion factor is discussed later in Sec.~\ref{sec:distortion}.

The view direction is the direction in which the an image is taken. Most of the common datasets for semantic segmentation, such as the Cityscapes, only include images taken in forward direction. When the view direction changes, the shapes, outlines and angles of the objects in the images will also change. So the problem is that if the model is trained only with images in forward direction, it's difficult for it to robustly segment panoramic images which consist of four direction images.

\begin{figure} [ht]
\centering
\subfigure[Segmentation result of model $M_1$] { \label{fig:forward}
\includegraphics[width=0.45\textwidth]{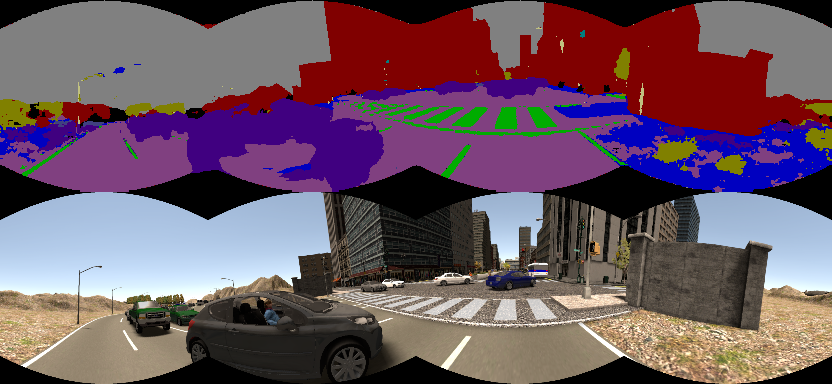}
}
\subfigure[Segmentation result of model $M_2$] { \label{fig:four_direction}
\includegraphics[width=0.45\textwidth]{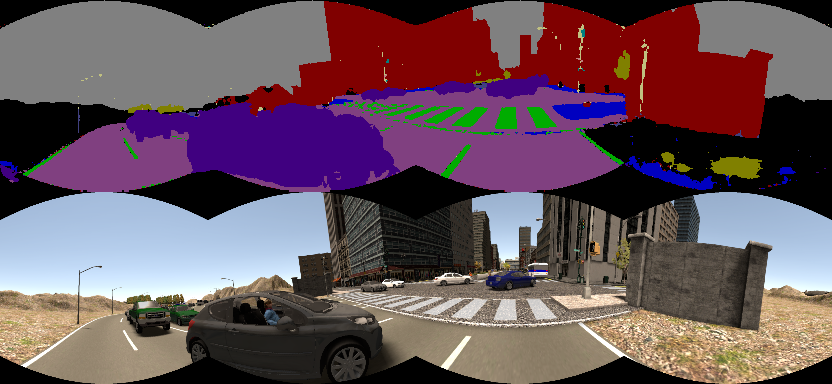}
}
\caption{These two figures show the results of two model trained on two sets of data which cover different views for comparison. (a) is of model $M_1$, which is trained with only forward direction images. (b) is of model $M_2$, which is trained with four direction images.}
\label{fig:view}
\end{figure}

We use a contrast experiment to illustrate this point. Specifically, we train the ICNet on two different datasets. The training data for model $M_1$ only include forward direction images while for model $M_2$ the data include four direction images.  The training images are all from SYNTHIA dataset (Seqs02-summer, Seqs02-fall, Seqs04-summer and Seqs04-fall) and the training settings are shown in Table.~\ref{tab:view_settings}. When training, the images from SYNTHIA dataset have been slightly resized to be exactly divisible by 16 in order to meet the down-sampling requirement of ICNet. Then we compare the performance of these two models on panoramic images in Seqs05-summer from SYNTHIA-PANO dataset and the segmentation results are shown in Fig.~\ref{fig:forward} and Fig.~\ref{fig:four_direction}.

\begin{table}[ht]
\centering
\caption{Training settings for models $M_1$, $M_2$}
\label{tab:view_settings}
\begin{tabular}{clcccc}
\hline
\multicolumn{1}{l}{Training settings} & image  size & \multicolumn{1}{l}{training set size} & \multicolumn{1}{l}{batch size} & \multicolumn{1}{l}{epoch} & \multicolumn{1}{l}{iteration} \\ \hline
$M_1$                                 & 1280, 768   & 1250                                  & 16                             & 200                       & 15k+                          \\
$M_2$                                 & 1280, 768   & 5000                                  & 16                             & 50                        & 15k+                          \\ \hline
\end{tabular}
\end{table}

Since the one direction training set does not include enough samples for all view conditions, in the area of left or right view, the result is not as fine as that in the area of front or back view. As for the segmentation results in the figures, it is clear that the car area which belongs to leftward direction view has been wrongly classified. Such ``collapse'' can be easily observed in the segmentation results of the model trained with only one direction images. This indicates that in order to obtain good segmentation results on panoramic images, it is necessary for the training set to cover different directions.

\subsection{Different FOVs for Training}
\label{sec:fov}

\begin{figure} [ht]
\centering
\subfigure[Training data of $M_{pano90}$] { \label{fig:pano90}
\includegraphics[width=0.25\textwidth]{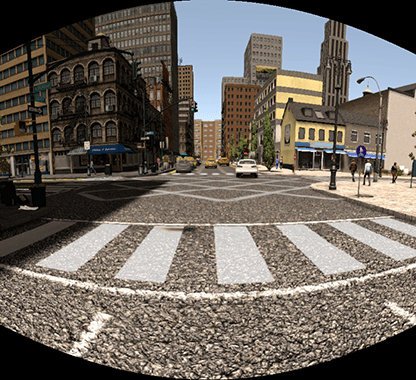}
}
\subfigure[Training data of $M_{pano180}$] { \label{fig:pano180}
\includegraphics[width=0.5\textwidth]{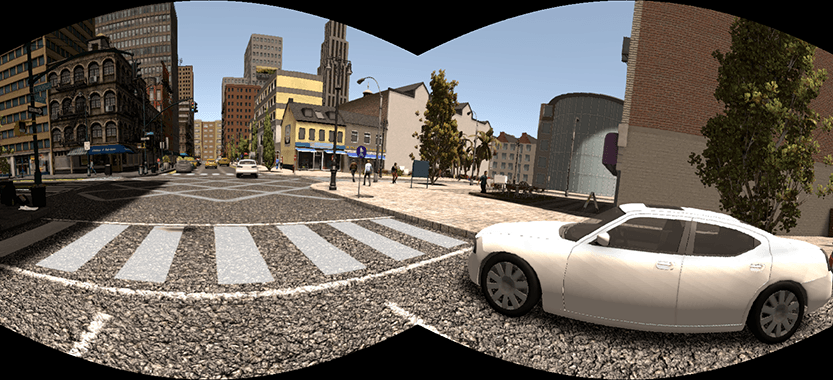}
}

\subfigure[Training data of $M_{pano360}$] { \label{fig:pano360}
\includegraphics[width=0.9\textwidth]{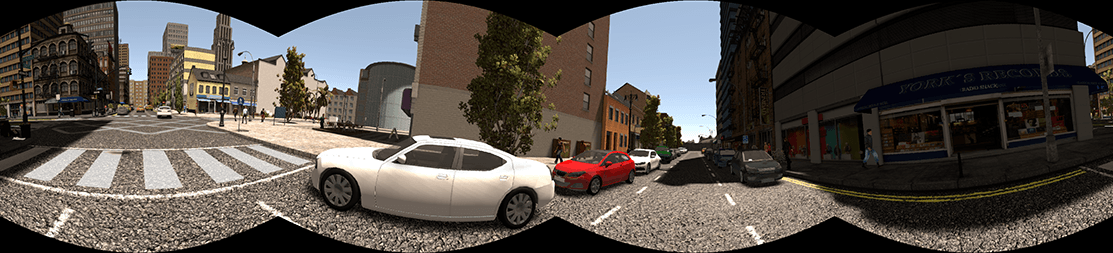}
}
\caption{ The figures show the training data of the models $M_{orin}$, $M_{pano90}$, $M_{pano180}$ and $M_{pano360}$. }
\label{fig:FOVmodels}
\end{figure}

In this subsection, an important setting is studied that is the FoV of the images for training. SYNTHIA-PANO is a panoramic image dataset and the images in it have FoVs of 360 degree. It's reasonable to presume that training with the large FoV of images should be beneficial for the panoramic segmentation result. Because a large FoV means that one image contains more information and has less discontinuity of the view. In this sense, the larger the FoV is, the better the result should be. However, the experiment show that the truth is not as expected.

We set up 4 contrasting models. Model $M_{orin}$ is the baseline model, which is trained with separated images of a 100 degree FoV in the original SYNTHIA dataset. Model $M_{pano90}, M_{pano180}, M_{pano360}$ are trained with the panoramic images in SYNTHIA-PANO dataset (Seqs02-summer, Seqs02-fall, Seqs04-summer and Seqs04-fall) and their FoVs are 90 degree, 180 degree and 360 degree, respectively. The training data for the four models are shown in Fig.~\ref{fig:FOVmodels}. The training settings are in Table.~\ref{tab:fov_settings}. When training, the images from SYNTHIA-PANO dataset have been slightly resized to be exactly divisible by 16 in order to meet the down-sampling requirement of ICNet. The relations among the batch size, epoch and the iteration need to be clarified. The total number of images of our training data is not fixed. Because for 1250 images of a 360 degree FoV, if we separate them into images of a 180 degree FoV, the number of images would become 2500. However, there is no difference in the contents of the images. To assure that the models converge at the same level and to ensure fair comparison, we fix the number of epochs and iterations. Moreover, in each batch, although the number of the input images is different for the models, they are actually fed with the same content.

\begin{table}[]
\centering
\caption{Training settings for models $M_{orin}$, $M_{pano90}$, $M_{pano180}$ and $M_{pano360}$}
\label{tab:fov_settings}
\begin{tabular}{@{}cccccc@{}}
\toprule
\multicolumn{1}{l}{Training settings} & \multicolumn{1}{l}{image  size} & \multicolumn{1}{l}{training set size} & \multicolumn{1}{l}{batch size} & \multicolumn{1}{l}{epoch} & \multicolumn{1}{l}{iteration} \\ \midrule
$M_{orin}$                            & 1280, 768                       & 5000                                  & 16                             & 200                       & 60k+                          \\
$M_{pano90}$                          & 832, 768                        & 5000                                  & 16                             & 200                       & 60k+                          \\
$M_{pano180}$                         & 1664, 768                       & 2500                                  & 8                              & 200                       & 60k+                          \\
$M_{pano360}$                         & 3328, 768                       & 1250                                  & 4                              & 200                       & 60k+                          \\ \bottomrule
\end{tabular}
\end{table}

\begin{figure} [ht]
\centering
\subfigure[The mIoU result of different FoV models] { \label{fig:fov_miou}
\includegraphics[width=0.45\textwidth]{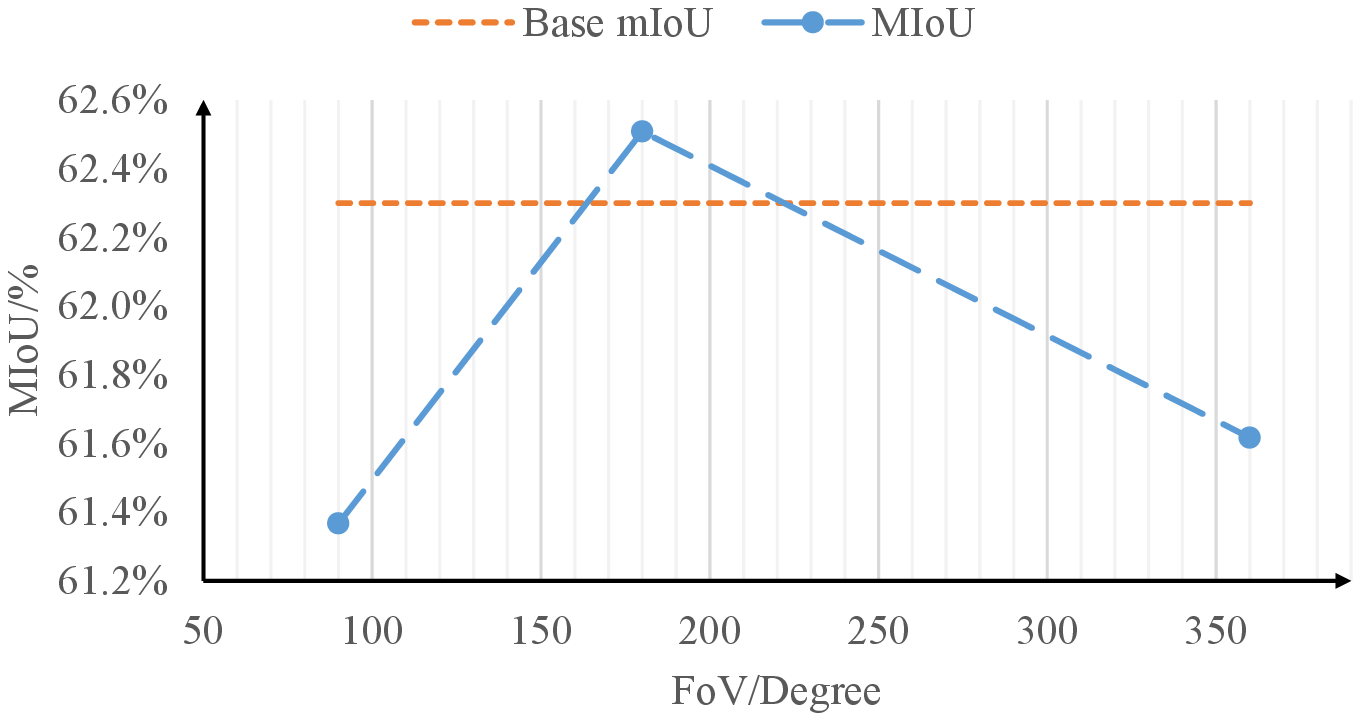}
}
\subfigure[The accuracy result of different FoV models] { \label{fig:fov_acc}
\includegraphics[width=0.45\textwidth]{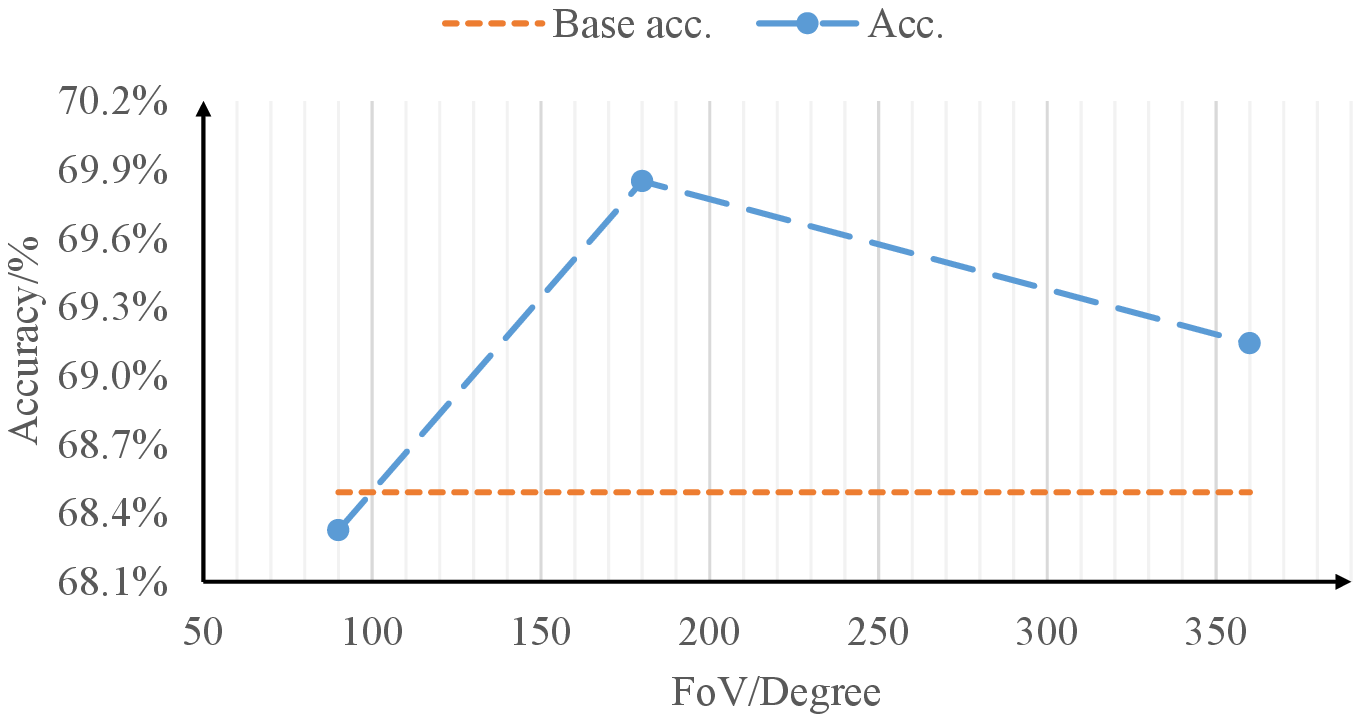}
}
\caption{The two figures show the mIoU and accuracy evaluation of the four models $M_{orin}$, $M_{pano90}$, $M_{pano180}$ and $M_{pano360}$.}
\label{fig:fov_result}
\end{figure}

All four models are evaluated on the Seqs05-summer in our SYNTHIA-PANO dataset and the results are shown in Table.~\ref{tab:fov_result} and Fig.~\ref{fig:fov_result}. The Fig.~\ref{fig:fov_miou} and Fig.~\ref{fig:fov_acc} give overall average evaluations while the Table.~\ref{tab:fov_result} shows the results of some specific classes of $M_{pano180}$ and $M_{orin}$ . As both of the two sub-figures in Fig.~\ref{fig:fov_result} shows when the FoV of training images increases, the accuracy and mIoU rise first and then fall down. The rising means the increase of the FoV can be beneficial to both the accuracy and mIoU, while the falling indicates that the too large FoV is not absolutely beneficial but may produce negative effect.

Its easy to understand a large FoV can boost the result. A small FoV not only breaks the spatial relations of the objects in the scene but also lacks consistency. For example, if only the head part of a car appears in the image with a small FoV, it will be more difficult to recognize it than in a large FoV image which contains all parts of the car.

It might be counter-intuitive that too larger FoV has a negative effect. So the reason why the too large FoV decreases the accuracy and mIoU needs to be further discussed. The panoramic images in SYNTHIA-PANO dataset have not only a large size but also an unbalanced aspect ratio which is nearly $3328:768\approx 4:1$. However, we use square kernels in the neural network model to extract and mix features. Such unmatched ratio may cause the loss of the features. In addition, the size of a panoramic image is often very large. The receptive field of ICNet may sufficiently cover image with 180-degree, but is not large enough when it comes to the whole 360-degree panorama. One solution is to make the model more complex by extending the network depth and enlarging the receptive field which will penalize the inference speed, so the better way is to choose images with a suitable FoV to train as it is demonstrated by our experiment.

The content above explains why a large FoV may bring bad effect, so it becomes necessary to find a relative balanced FoV which can avoid feature superfluity. According to the results of the experiment, a good selection is 180 degree. Although 360-degree FoV images contain much more information, some of the information are repeated and not well exploited. For instance, the scene in forward and leftward directions is quite like that in backward and rightward directions. In this sense, it's basically enough to include most of the useful features in a 180 FoV image.

\begin{table}[ht]
\centering
\caption{MIoU and accuracy of some specific classes benefited by training with 360-degree panoramic images}
\label{tab:fov_result}
\begin{tabular}{@{}lccccc@{}}
\toprule
Class             & \multicolumn{1}{l}{vegetation} & \multicolumn{1}{l}{car} & \multicolumn{1}{l}{traffic sign} & \multicolumn{1}{l}{pedestrian} & \multicolumn{1}{l}{traffic light} \\ \midrule
mIoU orin (\%)    & 64.4                           & 82.4                    & 26.1                             & 27.7                           & 33.5                              \\
Acc. orin (\%)    & 82.5                           & \textbf{91.7}           & 31.3                             & 33.3                           & 40.4                              \\
mIoU pano180 (\%) & \textbf{65.5}                  & \textbf{83.1}           & \textbf{27.0}                    & \textbf{30.0}                  & \textbf{34.0}                     \\
Acc. pano180 (\%) & \textbf{85.2}                  & 91.5                    & \textbf{31.6}                    & \textbf{37.9}                  & \textbf{42.3}                     \\ \bottomrule
\end{tabular}
\end{table}

For more detailed comparison in Table.~\ref{tab:fov_result}, the results of specific classes are discussed. Five critical classes are listed in Table.~\ref{tab:fov_result} , which shows the improvement using panoramic images as training data. Different from the large classes such as sky and building, these classes do not have large number of pixels but play important roles in the scene. After we stitch original images into panoramic images, the pixel numbers of these classes become even fewer, which is adverse for learning the features of these classes. However, the final results show the accuracy and mIoU of these classes are improved. The common property of these classes is that they have relatively fixed spatial positions. For example, the pedestrians and vegetation are generally on the sidewalk, while normally the cars, traffic signs and traffic lights are on the road. Such spatial relations can be observed more obviously in panoramic images than in normal images. This is because in panoramic images the patterns of a scene can be quite similar. A panoramic image is more likely to obtain more classes in it than an original image. The similarity of the scene in the images can be helpful for the model to recognize the spatial relations and to accurately segment the objects of these classes.


\subsection{Evaluation of Anti-distortion Ability}
\label{sec:distortion}

In Sec.~\ref{sec:view}, we discussed the view factor in panoramic images which may cause damage on the segmentation result. 
Another factor is the distortion of the panoramic images. 
We expect the model could be as robust as possible in not only slight distortion condition but also severe distortion condition. In order to find out whether training with panoramic images helps the model with it's anti-distortion ability, we perform another experiment.

\begin{figure} [ht]
\centering
\subfigure[$f=700$] { \label{fig:f700}
\includegraphics[width=0.3\textwidth]{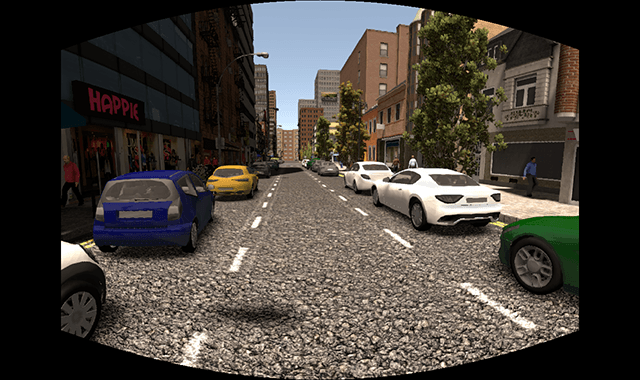}
}
\subfigure[$f=600$] { \label{fig:f600}
\includegraphics[width=0.3\textwidth]{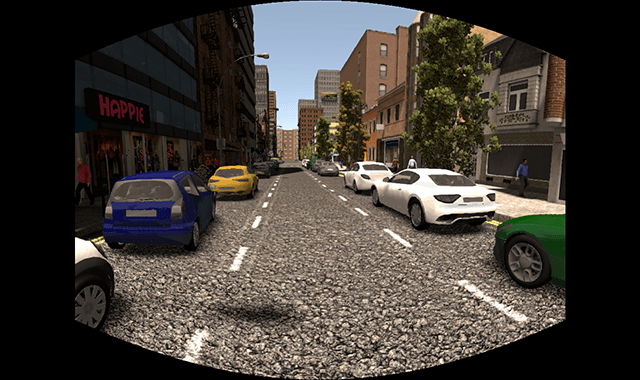}
}

\subfigure[$f=500$] { \label{fig:f500}
\includegraphics[width=0.3\textwidth]{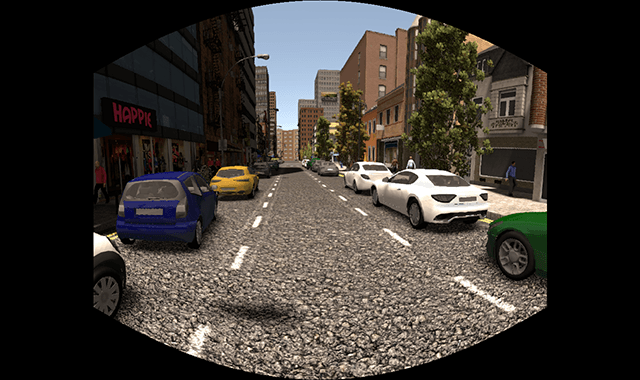}
}
\subfigure[$f=400$] { \label{fig:f400}
\includegraphics[width=0.3\textwidth]{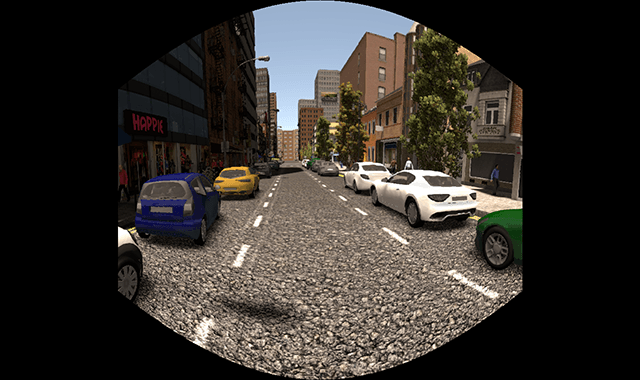}
}
\caption{ The four figures show images after cylindrical projection with different focal lengths $f$. In the experiment, $f$ is set to be $700, 600, 500,400$.}
\label{fig:focal}
\end{figure}

   \begin{figure} [ht]
   \begin{center}
   \begin{tabular}{c} 
   \includegraphics[width=0.6\textwidth]{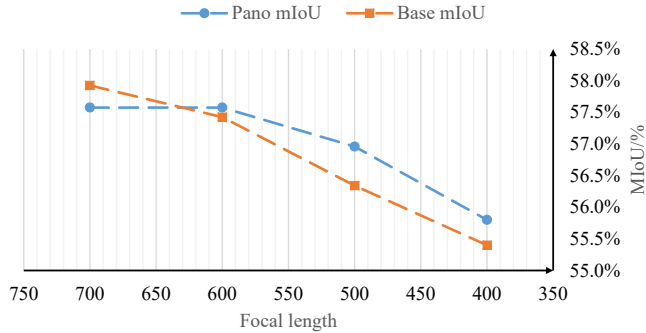}
   \end{tabular}
   \end{center}
   \caption[example]
   { \label{fig:distort}
The figure shows the mIoU of the two models $M_{orin}$ and $M_{pano180}$. When the focal length is in $700$ to $400$, the mIoU of $M_{orin}$ is higher than that of $M_{pano180}$.}
   \end{figure}

The models $M_{orin}$ and $M_{pano180}$ in  Sec.~\ref{sec:fov} are still used in the experiment. The model $M_{orin}$ is trained with original images in SYNTHIA dataset while the model $M_{pano180}$ is trained with panoramic images in SYNTHIA-PANO. In this experiment, the images used for evaluation have varying degrees of distortion, as it is shown in Fig.~\ref{fig:focal}. To obtain such groups of images, cylindrical projection is used and the focal length parameter is set to be a key variable, which varies from 700 to 400. The shorter the focal length is, the severer the distortion will be. We feed the two  models with the evaluation images and calculate the mIoU between the segmentation output results and the ground truth.

The results are in Fig.~\ref{fig:distort}. When the focal length is shorter than 625, $M_{orin}$ performs better than $M_{pano180}$. When distortion becomes more severe, both of the models' performance begin to drop. This indicates that the distortion of panoramic images does affect the segmentation result. However, it's obvious that the dropping rates of the two models are different. The dropping rate of $M_{orin}$ is approximately constant, which follows a linear decreasing fashion. But the dropping rate of $M_{pano180}$ is nearly 0 when focal length is longer than 600, and later it starts to increase. When the focal length is shorter than 625, the model $M_{pano180}$ trained with panoramic images always preforms better, which means that the anti-distortion ability of the model is strengthened. From this experiment, it is demonstrated the proposed training strategy using stitched panoramic images helps to attain robustness against severe distortions.

\section{CONCLUSION AND FUTURE WORK}

In order to obtain training data  for panorama segmentation, we propose an approach to stitch panoramic images and then synthesize a panoramic image dataset called SYNTHIA-PANO. Based on this dataset, we design and implement a set of experiments. 
First, the effect of the view of the training images is explored. One direction view tends to affect the segmentation result on panoramic images so that panoramic image training data is necessary. As for the FoV of the training images, 180 degree is proved to be a very suitable option. Last, the anti-distortion capacity of the different models is evaluated and the results indicate that the model trained with panoramic images has a better anti-distortion capacity. In summary, the model trained with panoramic images has the advantage of better performance and anti-distortion capacity to sense surroundings.

In practical applications, semantic segmentation of panoramic images plays an important role in fields such as autonomous driving, video surveillance and navigation assistance. It can expand the visual range, increase the invariability of scenes and ensure the continuity of semantic information. Based on our proposal, panoramic semantic segmentation can be addressed in a both efficient and robust way, which is immensely beneficial for these safety-critical applications.

Our further work may focus on the gap between synthetic data and real data. The synthetic data are easier to obtain and more information can be attached with the label images, for example, the depth information. So the synthesizing method is an important way which will be used more widely in the future. However, since synthetic dataset like the SYNTHIA dataset is typically generated by computer graphics, there are discrepancies between real images and synthetic images. As a result, if a model is trained with synthetic images, its performance will be limited when applied on real images. The gap between synthetic data and real data is the key point which limits the performance. We intend to fix the gap and help the synthetic data become more valuable.

\bibliography{report} 
\bibliographystyle{spiebib} 

\end{document}